\documentclass[a4paper,fleqn]{cas-sc}

\usepackage[authoryear,longnamesfirst]{natbib}

\def\tsc#1{\csdef{#1}{\textsc{\lowercase{#1}}\xspace}}
\tsc{WGM}
\tsc{QE}

\usepackage{tabularx}
\usepackage{multirow}
\usepackage{graphicx}
\usepackage{url}

\begin{document}
\let\WriteBookmarks\relax
\def\floatpagepagefraction{1}
\def\textpagefraction{.001}


\shortauthors{Xingliang Lei et al.}  

\title [mode = title]{Pre-training Everywhere: Parameter-Efficient Fine-Tuning for Medical Image Analysis via Target Parameter Pre-training}  

\tnotemark[1] 

\tnotetext[1]{This work was supported by the National Natural Science Foundation of China under Grants 92470101.} 

%
















\author[1]{Xingliang Lei}[orcid=0009-0003-6260-4061]
\fnmark[1]
\ead{leixingliang@mail.nwpu.edu.cn}

\author[1]{Yiwen Ye}[orcid=0000-0003-2189-6865]
\fnmark[1]
\ead{ywye@mail.nwpu.edu.cn}

\author[1]{Zhisong Wang}[orcid=0009-0009-4764-3699]
\ead{zswang@mail.nwpu.edu.cn}

\author[1]{Ziyang Chen}[orcid=0000-0002-8564-9735]
\ead{zychen@mail.nwpu.edu.cn}

\author[2]{Minglei Shu}[orcid=0000-0002-7136-1538]
\ead{shuml@sdas.org}

\author[3]{Weidong Cai}[orcid=0000-0003-3706-8896]
\ead{tom.cai@sydney.edu.au}

\author[1]{Yanning Zhang}[orcid=0000-0002-2977-8057]
\ead{ynzhang@nwpu.edu.cn}

\author[1]{Yong Xia}[orcid=0000-0001-9273-2847]
\cormark[1]
\ead{yxia@nwpu.edu.cn}

\fntext[1]{These authors contributed equally to this work.}
\cortext[cor1]{Corresponding author.}

\affiliation[1]{organization={School of Computer Science and Engineering, Northwestern Polytechnical University},
            city={Xi'an},
            postcode={710072}, 
            state={Shaanxi},
            country={China}}

\affiliation[2]{organization={Shandong Artificial Intelligence Institute, Qilu University of Technology},
            city={Jinan},
            postcode={250353}, 
            state={Shandong},
            country={China}}

\affiliation[3]{organization={School of Computer Science, The University of Sydney},
            city={Sydney},
            postcode={2006}, 
            state={New South Wales},
            country={Australia}}


\begin{abstract}
Parameter-efficient fine-tuning (PEFT) techniques have emerged to address overfitting and high computational costs associated with fully fine-tuning in self-supervised learning. Mainstream PEFT methods add a few trainable parameters while keeping the pre-trained backbone parameters fixed. These methods achieve comparative, and often superior, performance to fully fine-tuning, demonstrating the powerful representation ability of the pre-trained backbone. 
Despite this success, these methods typically ignore the initialization of the new parameters, often relying solely on random initialization. We argue that if pre-training is significantly beneficial, it should be applied to all parameters requiring representational capacity. Motivated by this, we propose Target Parameter Pre-training (TPP), a simple yet effective fine-tuning framework. 
TPP pre-trains target parameters, \textit{i.e.}, the new parameters introduced during fine-tuning, in an additional stage before PEFT. During this stage, the pre-trained backbone parameters are frozen, and only the new parameters are trainable. A defined pretext task encourages the new parameters to learn specific representations of downstream data. Subsequently, when PEFT is employed, the pre-trained new parameters are loaded to enhance fine-tuning efficiency. The proposed TPP framework is versatile, allowing integration with various pre-trained backbones, pretext tasks, and PEFT methods. 
We evaluated the fine-tuning performance of our method on seven public datasets, covering four modalities and two task types. The results demonstrate that TPP can be easily integrated into existing PEFT methods, significantly improving performance.
\end{abstract}








\maketitle

\begin{figure}[!ht]
\centerline{\includegraphics[width=0.6\linewidth]{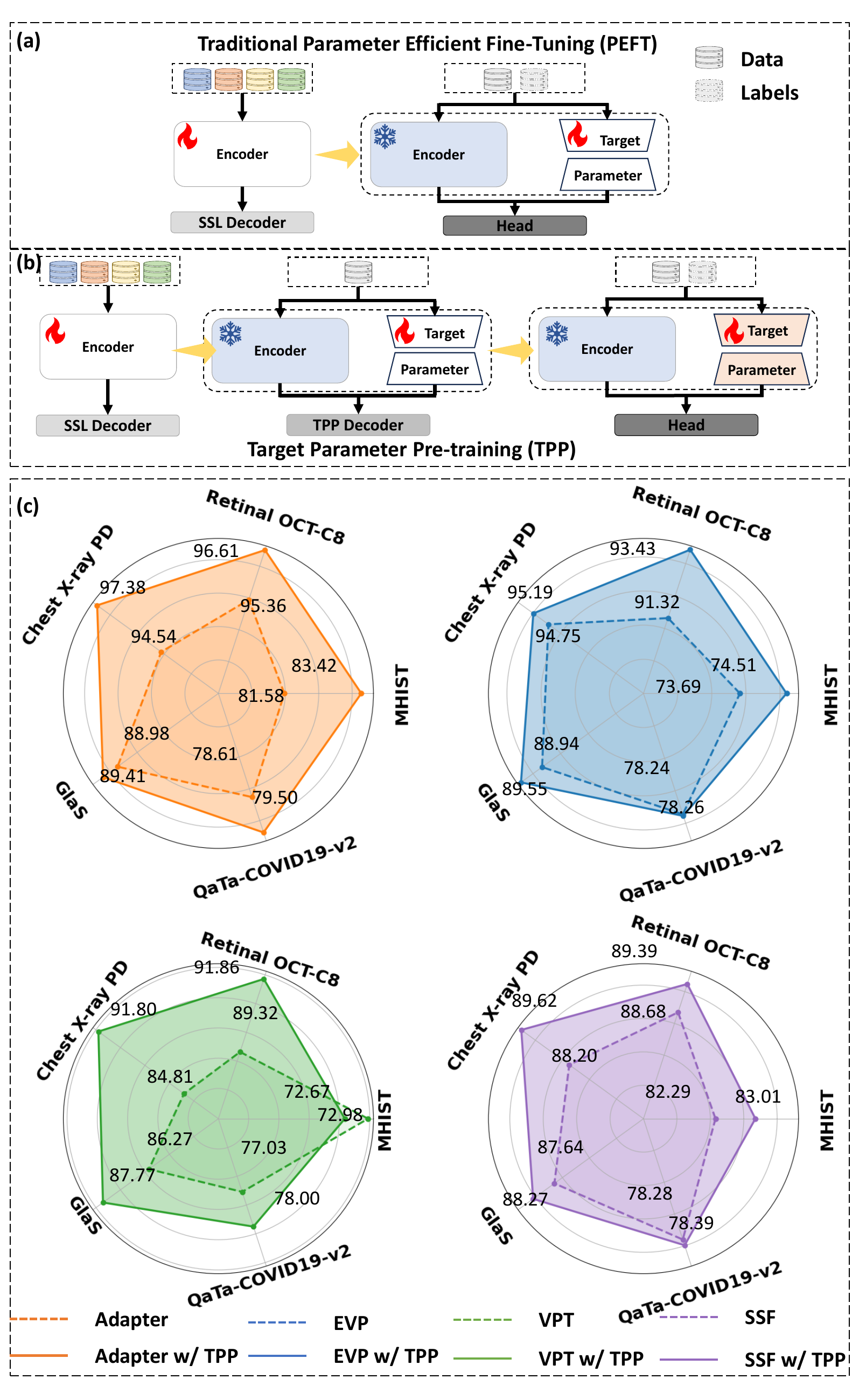}}
\caption{Comparison between (a) Traditional SSL paradigm, \textit{i.e.}, pre-training then fine-tuning, and (b) Our TPP framework. The newly introduced stage is inserted between the pre-training stage and fine-tuning stage and is used to pre-train the target parameters. (c) Performance comparison of PEFT methods and these methods with TPP based on MAE on MedCoSS, represented by using dashed and solid lines, respectively. The different colors represent different PEFT methods.}
\label{fig:figure1}
\end{figure}

\section{Introduction}
\label{sec:introduction}
Deep learning has significantly advanced medical image analysis, markedly improving model performance \cite{unet++, greenspan2016guest}. However, these models often require large-scale, high-quality annotated datasets, which are particularly time-consuming and labor-intensive to produce in medical imaging \cite{parameter, wu2023coactseg}. Consequently, medical image analysis frequently encounters challenges related to limited dataset sizes \cite{willemink2020preparing}, which hinders its clinical application.
Self-supervised learning (SSL) has gained prominence due to its reduced data demands, high accuracy, and strong generalization ability. SSL typically involves two stages: pre-training on a large-scale unlabeled dataset and fine-tuning on labeled datasets. While various pretext tasks are employed for pre-training, SSL-pre-trained models \cite{SimCLR, azizi2021big, path_dino, MedCoSS} are typically fully fine-tuned during the downstream task, updating all pre-trained parameters. This approach can be problematic for medical imaging tasks with limited fine-tuning data, as updating all pre-trained parameters on a small dataset can compromise model robustness and elevate the risk of overfitting \cite{parameter}. 

To address these challenges, parameter-efficient fine-tuning (PEFT) has emerged as a promising solution \cite{parameter}. 
PEFT typically involves freezing pre-trained parameters and either updating a subset of these parameters \cite{bitfit, attention} or adding a small set of new trainable parameters, known as \textit{target parameters} \cite{lora, adapter, adaptformer, dtl, lst, vpt, prefix, spt, dvpt} (see Figure \ref{fig:figure1}(a)). Methods that add target parameters have demonstrated superior performance while utilize only a small number of new parameters. However, despite leveraging the robust representations of pre-trained models, these methods do not fully exploit the benefits of pre-training. We argue that \textit{if the backbone is pre-trained, the newly added parameters should also undergo pre-training}.

In this paper, we propose Target Parameter Pre-training (TPP), a simple yet effective fine-tuning framework. Our TPP framework focuses on pre-training the new parameters introduced by PEFT methods, excluding task-specific heads (\textit{i.e.}, linear layers for classification or decoders for segmentation), which we refer to as target parameters. Unlike existing PEFT strategies that directly fine-tune pre-trained models with randomly initialized target parameters, our approach introduces a target parameter pre-training stage between pre-training and fine-tuning (see Figure \ref{fig:figure1}(b)). 
This stage involves pre-training the target parameters by freezing the well-pre-trained backbone and updating the target parameters through a pretext task. Specifically, we pre-train the target parameters for each downstream dataset using its training data, enabling them to learn dataset-specific representations. 
Simultaneously, by integrating new knowledge into the target parameters, our approach enhances the overall representational capacity of the model.
We evaluated our TPP framework on seven public datasets, comparing its performance with seven PEFT methods, full fine-tuning, and linear probe. As a plug-and-play solution, TPP can be integrated into pre-trained backbones and consistently improves existing PEFT methods (see Figure \ref{fig:figure1}(c)).

The contributions of this work are three-fold.
\begin{itemize}
\item We propose TPP, a plug-and-play PEFT framework for medical image classification and segmentation tasks. This framework can be easily integrated into existing PEFT methods that add target parameters without requiring architectural modifications.
\item We introduce the insight that if the backbone is pre-trained, the target parameters should also be pre-trained, and therefore propose a TPP stage that optimizes target parameters for specific downstream data.
\item We comprehensively demonstrate that the proposed TPP framework consistently enhances PEFT baselines across seven datasets, covering two task types and four imaging modalities.
\end{itemize}

\section{Related Work}
\subsection{PEFT Techniques}
PEFT techniques aim to minimize the number of updated parameters. An initial approach involves updating specific pre-trained model parameters, such as biases \cite{bitfit}, layer normalization \cite{layernorm}, batch normalization \cite{batchnorm}, and attention mechanisms \cite{attention}. However, this can disrupt the pre-trained knowledge, potentially leading to suboptimal performance.

To address this issue, recent research has focused on introducing new trainable parameters, termed target parameters. In this strategy, the parameters of the pre-trained backbone are kept fixed while the target parameters are updated using downstream datasets. This strategy can be broadly categorized into four main approaches.
(1) \textbf{Adapter-based methods:} These methods \cite{adapter, adaptformer, convadapter, med-tuning} insert additional blocks in parallel or serially to learn residual features. For instance, Adapter \cite{adapter} inserts a bottleneck block with residual connections serially between the feed-forward layer of Transformer block. AdaptFormer \cite{adaptformer} employs a similar bottleneck structure without residual connections, placed in parallel within the MLP section of Transformer blocks.
(2) \textbf{Low-Rank Adaptation-based (LoRA) methods:} These methods \cite{lora, adalora, convlora, lisa} introduce learnable low-rank matrices into the self-attention layers of Transformers. LoRA \cite{lora} offers high parameter efficiency and integrates seamlessly with the pre-trained backbone during inference, minimizing additional computational overhead. Variants such as ConvLoRA \cite{convlora} and AdaLoRA \cite{adalora} enhance LoRA by incorporating lightweight convolutional parameters or adaptively allocating parameter budgets based on importance scores.
(3) \textbf{Prompting-based methods:} These methods \cite{vpt, dvpt, gatevpt, spt} modify the input to Transformer layers by adding learnable tokens. For example, VPT \cite{vpt} and its variants \cite{dvpt, gatevpt} introduce learnable tokens to the inputs of Transformer layers for visual tasks. Although parameter-efficient, these methods can be sensitive to prompt token lengths and may exhibit suboptimal performance when applied to self-supervised pre-trained backbones.
(4) \textbf{Side-tuning-based methods:} These methods \cite{fpt, dtl, yin2024parameter, last} utilize a separate side model that operates independently of the frozen backbone, avoiding gradient backpropagation through the backbone. For instance, DTL \cite{dtl} uses a lightweight Compact Side Network (CSN) with low-rank linear mappings to capture task-specific features, which are then integrated into the backbone.

\begin{figure*}[t]
    \centering
    \includegraphics[width=\linewidth]{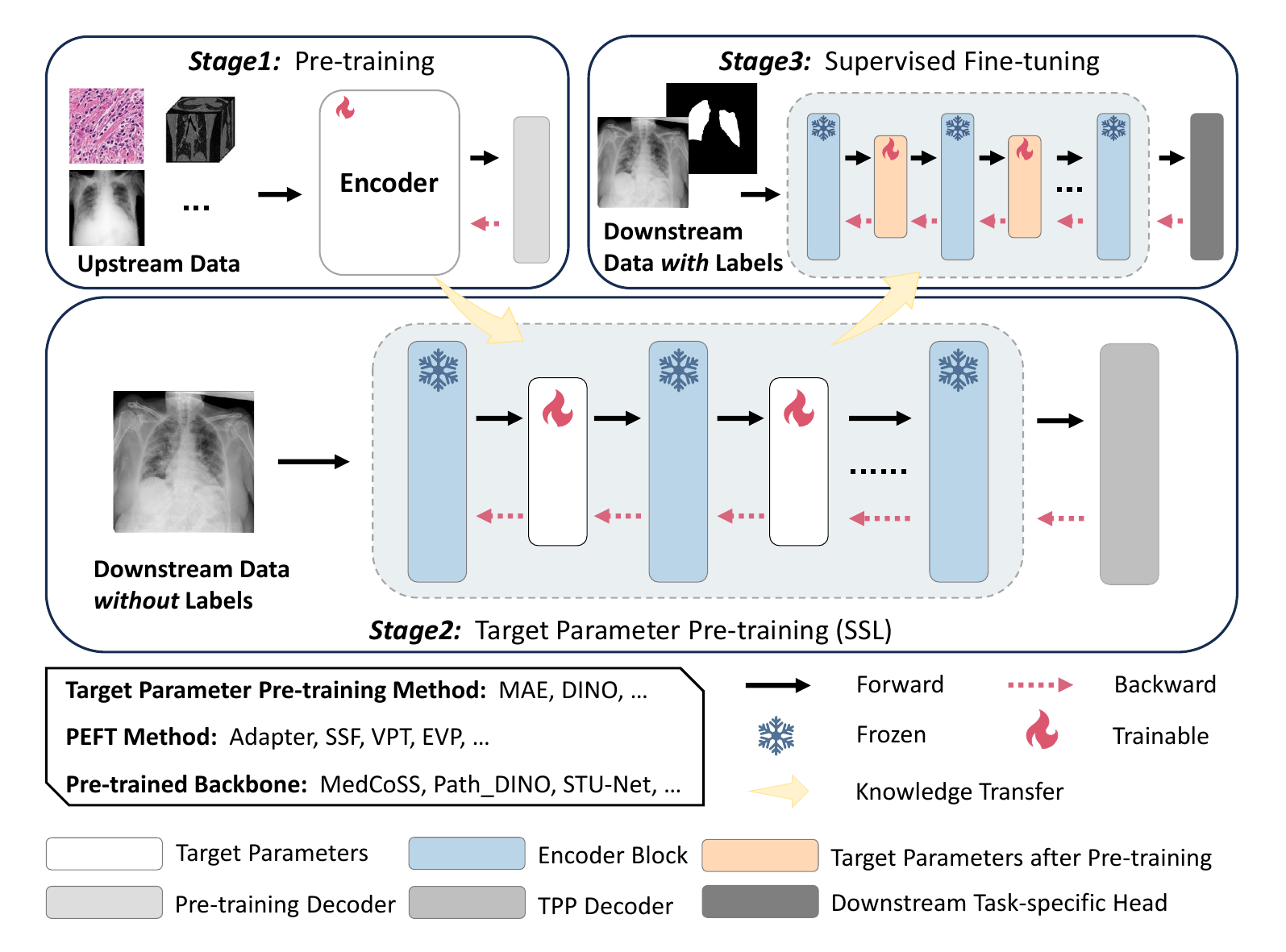} 
    \caption{Overview of the proposed TPP framework. It introduces a pre-training stage for target parameters between the traditional pre-training and the subsequent supervised fine-tuning stages.
    }
    \label{fig:overrall}
\end{figure*}

Despite the progress in PEFT, many existing methods initialize target parameters randomly (\textit{e.g.}, using zero, uniform distributions, or normal distributions) \cite{adapter, lora, ssf, vpt}. For instance, LoRA typically initializes its two low-rank matrices with zero and random values, respectively. These methods aim to ensure that target parameters do not alter the input to the original layer initially \cite{liao2024make}. However, this approach overlooks the potential benefits of pre-training for these new parameters. Given the established advantages of pre-training for backbone parameters, it is reasonable to expect similar benefits for target parameters.

Recent efforts have explored improving target parameter initialization. Self-Prompt Tuning \cite{spt} initializes prompts using token prototypes derived from target data, specifically for prompting-based methods. 
PVP \cite{pvp} pre-trains target parameters on a large-scale dataset, and Su \textit{et al.} \cite{su2022transferability} initializes target parameters with soft prompts pre-trained on related tasks. 
However, these methods do not fully address the challenge of selecting appropriate datasets for pre-training target parameters. We argue that pre-training directly on the target dataset offers a more practical and effective solution. Therefore, our proposed TPP framework pre-trains target parameters using the current target training data to enhance subsequent fine-tuning performance.

Cekmeceli \textit{et al.} \cite{cekmeceli2024vision} investigated the domain generalization performance of foundation models fine-tuned with various PEFT methods, emphasizing the importance of selecting appropriate techniques to ensure that target parameters trained on source domain data can be directly applied to target domain data. In contrast, our work focuses on the broad impact of target parameter initialization on performance in target tasks.

\subsection{SSL Methods}
SSL is a powerful technique for representation learning \cite{SimCLR, moco, dino, autoencoder, dae, mae}. SSL typically consists of two stages: unsupervised pre-training on a large scale unlabeled data with a pretext task and supervised fine-tuning on target data.

Various pretext tasks have been explored, including contrastive learning \cite{SimCLR, moco, dino} and generative learning \cite{autoencoder, dae, mae}, as well as combinations thereof \cite{gan}.
For instance, Masked Autoencoders (MAE) \cite{mae} employs masked image modeling, where an image is divided into patches, most of which are randomly masked. The model learns to predict these masked patches from the remaining visible ones. DINO \cite{dino} combines multi-crop learning \cite{caron2020unsupervised} with self-distillation for contrastive learning, utilizing centering and sharpening operations to prevent collapse in the momentum teacher model's output vectors.
SSL pre-training enables models to learn high-quality representations, reducing the reliance on large annotated datasets and accelerating convergence. This is particularly valuable for various medical image analysis tasks, including classification \cite{zhou2021preservational, MedCoSS}, localization \cite{yang2021instance, simeoni2021localizing}, and segmentation \cite{zhou2021preservational, MedCoSS, ye2022desd}.

Despite the benefits of SSL for pre-training backbones, PEFT methods often initialize new parameters randomly, potentially undermining these benefits. To address this, our study introduces an additional pre-training stage for these new parameters before fine-tuning, aiming to fully leverage the potential of SSL.

\section{Method}
\subsection{Preliminaries}
\noindent{\textbf{Paradigm of SSL.}}
SSL typically involves two stages: pre-training and fine-tuning.
Let $D_{un}$ denote a large-scale unlabeled dataset, and $\theta$ represent the parameters of the backbone network.
In the pre-training stage, $\theta$ is trained on $D_{un}$ by completing a pretext task that does not require human-labeled data for supervision. The parameters obtained after pre-training, denoted as $\theta^{p}$, exhibit strong transferability to various downstream tasks.
In the fine-tuning stage, a labeled dataset $D_{l}$ is used to adapt the pre-trained backbone to the target task. This is achieved by adding a new task-specific head, represented by $\theta_{head}$, and updating all parameters. The process of a single feed-forward operation is described as:
\begin{equation}
P = f(I, \langle\theta^{p}, \theta_{head}\rangle), I \in D_{l},
\end{equation}
where $I$ is the input image, $\langle , \rangle$ denotes the set of parameters involved in the feed-forward process $f(,)$, and $P$ is the output.

\noindent{\textbf{Pre-trained Backbone.}}
We primarily employ MedCoSS \cite{MedCoSS} as the pre-trained backbone, due to its large-scale multi-modal pre-training and robust generalization ability across downstream tasks with various modalities. MedCoSS is based on ViT/B \cite{vit} and pre-trained using the pretext task of masked modeling \cite{devlin2019bert, mae}. To demonstrate the general applicability of TPP, we also experimented with two additional backbones: Path\_DINO \cite{path_dino}, which is pre-trained on pathological images using DINO \cite{dino}, and STU-Net \cite{stunet}, a CNN-based medical foundation model pre-trained on the TotalSegmentator dataset \cite{totalsegmentator} in a supervised learning manner.

\noindent{\textbf{PEFT based on Adding New Parameters.}}
Leveraging the representation capabilities of pre-trained models, effective PEFT can be achieved by freezing the pre-trained model parameters and introducing new trainable parameters, termed target parameters $\theta_{tp}$. The feed-forward process can then be described as:
\begin{equation}
P = f(I, \langle|\theta^{p}|, \theta_{tp}, \theta_{head}\rangle), I \in D_{l},
\end{equation}
where $|\cdot|$ indicates the frozen parameters.

\subsection{Target Parameter Pre-training}
To address the limitations of randomly initialized target parameters, which may not fully exploit the benefits of pre-training, we introduce a TPP stage between the self-supervised pre-training and fully-supervised fine-tuning,  as depicted in Figure \ref{fig:overrall}. During TPP, we freeze the pre-trained backbone parameters $\theta^{p}$, while keeping the target parameters $\theta_{tp}$ trainable. Given a target dataset $D_{l}$, we pre-train $\theta_{tp}$ by performing a defined pretext task. We explored two popular pretext tasks, MAE \cite{mae} and DINO \cite{dino}, to validate the effectiveness of TPP. Empirically, TPP based on MAE generally outperforms TPP based on DINO across most PEFT methods.
We now delve in the details.

\noindent{\textbf{TPP based on MAE.}}
MAE is a generative SSL pretext task that divides an image into multiple non-overlapping patches, randomly masks a significant portion (typically 75\%), and uses the visible patches to predict the masked ones. A mean squared error (MSE) loss function is used to minimize prediction errors. This task exploits the redundancy in images, forcing the model to learn rich contextual representations that capture both local and global features.

\noindent{\textbf{TPP based on DINO.}}
DINO is a contrastive SSL pretext task. For an image,  strong data augmentations are used to generate two distinct views, which are processed by both the student and teacher models. The student model is a pre-training backbone followed by multi-layer perceptrons (MLPs), while the teacher model is a momentum-updated version of the student model. The student model is trained to maximize the similarity of the output vectors to those of the teacher model using a cross-entropy loss function.

Using both SSL pretext tasks, we obtain the pre-trained target parameters $\theta_{tp}^{p}$. These parameters, along with $\theta^{p}$, are then transferred to the target dataset $D_{l}$. Following PEFT methods, we freeze $\theta^{p}$ and keep  $\theta_{tp}^{p}$ and $\theta_{head}$ trainable. The feed-forward process is defined as:
\begin{equation}
P = f(I, \langle|\theta^{p}|, \theta_{tp}^{p}, \theta_{head}\rangle), I \in D_{l}.
\end{equation}
Our TPP framework is designed to be a plug-and-play solution, seamlessly integrating with any PEFT method that utilizes target parameters without requiring architectural modifications.

\begin{table}[t]
\centering
\small
\caption{Overview of the datasets used in this study.}
\label{tab: dataset}
\begin{tabular}{llllll}
\hline
\multicolumn{1}{l}{Dataset} & Modality & Task & Train & Val & Test \\ \hline
MHIST                        & Path.    & Cls       & 1740  & 435          & 977    \\
Retinal OCT-C8               & OCT      & Cls       & 18400 & 2800       & 2800 \\
Chest X-ray PA		 & X-ray    & Cls       & 3201  & 459          & 915    \\
GlaS                         & Path.    & Seg       & 67    & 18          & 80   \\
QaTa                   & X-ray    & Seg       & 5716  & 1429          & 2113 \\\hline
\end{tabular}
\end{table}

\begin{table*}[t!]
\centering
\small
\caption{Implementation details of five downstream tasks. CE: cross-entropy loss function.}
\label{tab:imple_Detail}
\begin{tabularx}{\textwidth}{c *{1}{>{\centering\arraybackslash}X} *{1}{>{\centering\arraybackslash}X} *{1}{>{\centering\arraybackslash}X} *{1}{>{\centering\arraybackslash}X} *{1}{>{\centering\arraybackslash}X}}
\hline
Dataset                           & Loss Function & Patch Size & Learning Rate & Batch Size & Iterations \\ \hline
MHIST                             & CE                     & $224\times224$             & Grid Search              & 32                  & 2,700                \\
Retinal OCT-C8                    & CE                     & $224\times224$             & 0.0001                   & 32                  & 57,500               \\
Chest X-ray PA & CE                     & $224\times224$             & 0.0001                  & 32                  & 1,980                \\
GlaS                               & Dice+CE                & $512\times512$             & 0.0001                   & 4                   & 25,000                \\
QaTa                         & Dice+CE                & $224\times224$             & 0.0001                   & 16                  & 25,000                \\ \hline
\end{tabularx}
\end{table*}  

\begin{table*}[t!]
\centering
\small
\caption{Results of Adapter with TPP, full fine-tuning, linear probe, and seven PEFT methods on three classification downstream tasks. The best result on each column is highlighted in \textbf{bold}. Ratio denotes the ratio of trainable parameters to total parameters.}
\label{tab:cls}
\begin{tabularx}{\linewidth}{c *{4}{>{\centering\arraybackslash}X}|*{4}{>{\centering\arraybackslash}X} |*{4}{>{\centering\arraybackslash}X}}
\hline
\multirow{2}{*}{Method} & \multicolumn{4}{c|}{MHIST} & \multicolumn{4}{c|}{Retinal OCT-C8} & \multicolumn{4}{c}{Chest X-ray PA} \\ \cline{2-13} 
                                 & ACC & AUC & F1 &Ratio & ACC & AUC & F1 &Ratio & ACC & AUC & F1 &Ratio \\ \hline
full fine-tuning & 83.11          & 89.78          & 81.60          & 100.0 & 94.54          & 99.64          & 94.53          & 100.0 & 93.33 & 98.97 & 93.34 & 100.0 \\
linear probe    & 68.99          & 77.08          & 66.46          & 0.003   & 59.96          & 92.64          & 59.42          & 0.008   & 74.97          & 97.45          & 70.03          & 0.004   \\
EVP              & 73.69          & 81.34          & 71.82          & 0.961   & 91.32          & 99.34          & 91.31          & 0.966   & 94.75          & 99.21          & 94.72          & 0.962   \\
VPT            & 72.98          & 79.84          & 71.35          & 0.103   & 89.32          & 99.04          & 89.25          & 0.108   & 84.81          & 95.33          & 84.58          & 0.103   \\
SSF             & 82.29          & 89.29          & 80.85          & 0.172   & 88.68          & 98.93          & 88.66          & 0.177   & 88.20          & 97.78          & 87.81          & 0.173   \\
BitFit          & 82.40          & 89.60          & 80.86          & 0.115   & 87.18          & 98.81          & 87.13          & 0.120   & 87.65          & 98.19          & 87.18          & 0.116   \\
Adapter         & 81.58          & 89.09          & 80.16          & 3.045   & 95.36          & 99.74          & 95.36          & 3.050   & 94.54          & 99.50          & 94.48          & 3.046   \\
AdaptFormer      & 66.02          & 74.30          & 65.07          & 3.831   & 82.32          & 98.11          & 82.37          & 3.836   & 79.23          & 92.77          & 78.95          & 3.832   \\
DTL              & 77.48          & 84.85          & 75.09          & 0.051   & 89.79                & 99.13                & 89.81                & 0.055   & 90.71          & 98.40          & 90.51          & 0.051   \\
Adapter w/ TPP            & \textbf{83.42} & \textbf{91.38} & \textbf{82.21} & 3.045   & \textbf{96.61} & \textbf{99.77} & \textbf{96.61} & 3.050   & \textbf{97.38}          & \textbf{99.63}          & \textbf{97.37}          & 3.046  
                               \\ \hline
\end{tabularx}
\end{table*}

\begin{table}[htbp]
\centering
\small
\caption{Results of Adapter with TPP, full fine-tuning, linear probe, and seven PEFT methods on two segmentation downstream tasks. The best result on each column is highlighted in \textbf{bold}. Ratio denotes the ratio of trainable parameters to total parameters.}
\label{tab:seg}
\begin{tabularx}{0.6\linewidth}{c *{3}{>{\centering\arraybackslash}X} @{\hspace{10pt}} |*{3}{>{\centering\arraybackslash}X}}
\hline
\multirow{2}{*}{Method} & \multicolumn{3}{c|}{GlaS}                & \multicolumn{3}{c}{QaTa}                \\ \cline{2-7} 
                                 & Dice  & HD    & Ratio & Dice  & HD    & Ratio \\ \hline
full fine-tuning        & 88.84          & 47.41          & 100.0          & 78.63          & 29.54          & 100.0          \\
linear probe            & 86.56          & 61.85          & 2.000          & 73.62          & 39.46          & 2.000          \\
EVP                    & 88.94          & 49.10          & 2.940          & 78.24          & 31.67          & 2.940          \\
VPT                     & 86.27          & 52.23          & 2.100          & 77.03          & 33.43          & 2.100          \\
SSF                    & 87.64          & 54.49          & 2.170          & 78.28          & 30.53          & 2.170          \\
BitFit                  & 87.56          & 58.58          & 2.110          & 77.99          & 32.30          & 2.110          \\
Adapter                & 88.98          & 49.06          & 4.980          & 78.61          & 29.91          & 4.980          \\
AdaptFormer             & 86.62          & 61.43          & 5.750          & 74.97          & 36.23          & 5.750          \\
DTL                     & 87.52          & 55.11          & 2.047          & 73.99          & 38.29          & 2.047          \\
Adapter w/ TPP                    & \textbf{89.41} & \textbf{44.25} & 4.980          & \textbf{79.50} & \textbf{29.28} & 4.980          \\ \hline
\end{tabularx}
\end{table}

\begin{figure*}[t]
    \centering
    \includegraphics[width=0.95\linewidth]{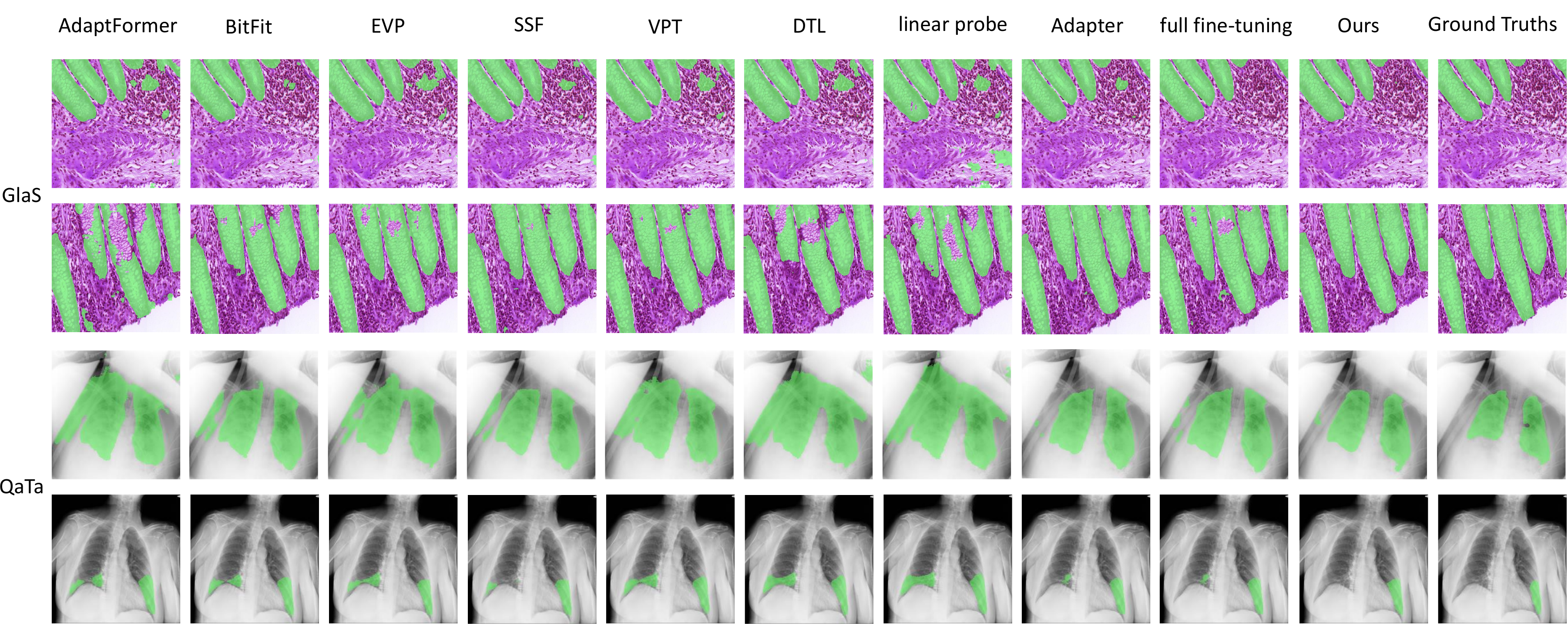} 
    \caption{Visualization of segmentation results obtained by AdaptFormer, BitFit, EVP, SSF, VPT, DTL, linear probe, Adapter, full fine-tuning, Ours, and Ground Truths. The pneumonia and malignant regions are colored green.}
    \label{fig:visual}
\end{figure*}

\begin{table*}[t]
\centering
\small
\caption{Results on four PEFT Methods based on TPP. For TPP, we used MAE as the pretext task. We highlight in \textbf{bold} the better results comparing each PEFT method with and without TPP.}
\label{tab:PEFT}
\begin{tabularx}{\linewidth}{c *{3}{>{\centering\arraybackslash}X} |*{3}{>{\centering\arraybackslash}X} |*{3}{>{\centering\arraybackslash}X} |*{2}{>{\centering\arraybackslash}X} |*{2}{>{\centering\arraybackslash}X}}
\hline
\multicolumn{1}{c}{\multirow{2}{*}{Method}} & \multicolumn{3}{c|}{MHIST}        & \multicolumn{3}{c|}{Retinal OCT-C8}          & \multicolumn{3}{c|}{Chest X-ray PA} & \multicolumn{2}{c|}{GlaS} & \multicolumn{2}{c}{QaTa} \\ \cline{2-14} 
\multicolumn{1}{c}{}                                 & ACC & AUC & F1 & ACC & AUC & F1 & ACC & AUC & F1 & Dice    & HD    & Dice    & HD    \\ \hline
Adapter     & 81.58 & 89.09 & 80.16 & 95.36 & 99.74 & 95.36 & 94.54 & 99.50 & 94.48 & 88.98 & 49.06 & 78.61 & 29.91 \\
Adapter w/ TPP &\textbf{83.42} & \textbf{91.38} & \textbf{82.21} & \textbf{96.61} & \textbf{99.77} & \textbf{96.61} & \textbf{97.38} & \textbf{99.63} & \textbf{97.37} & \textbf{89.41} & \textbf{44.25} & \textbf{79.50} & \textbf{29.28} \\
\hline
EVP         & 73.69 & 81.34 & 71.82 & 91.32 & 99.34 & 91.31 & 94.75 & 99.21 & 94.72 & 88.94 & 49.10 & 78.24 & 31.67 \\ 
EVP w/ TPP      & \textbf{74.51} & \textbf{82.17} & \textbf{72.13} &\textbf{93.43} & \textbf{99.47} &\textbf{93.43} & \textbf{95.19} & \textbf{99.36} & \textbf{95.18} & \textbf{89.55} & \textbf{48.06} & \textbf{78.26} & \textbf{31.23} \\
\hline
VPT          & \textbf{72.98} & 79.84 & 71.35 & 89.32 & 99.04 & 89.25 & 84.81 & 95.33 & 84.58 & 86.27 & \textbf{52.23} & 77.03 & 33.43 \\
VPT  w/ TPP      & 72.67 & \textbf{80.19} & \textbf{71.55} & \textbf{91.86} & \textbf{99.28} & \textbf{91.85} & \textbf{91.80} & \textbf{98.71} & \textbf{91.66} & \textbf{87.77} & 57.06 & \textbf{78.00} & \textbf{31.98} \\
\hline
SSF    & 82.29 & 89.29 & 80.85 & 88.68 & 98.93 & 88.66 & 88.20 & 97.78 & 87.81 & 87.64 & \textbf{54.49} & 78.28 & \textbf{30.53} \\
SSF  w/ TPP      & \textbf{83.01} & \textbf{90.50} & \textbf{81.72} & \textbf{89.39} & \textbf{99.05} & \textbf{89.35} & \textbf{89.62} & \textbf{98.44} & \textbf{89.26} & \textbf{88.27} & 57.36 & \textbf{78.39} & 31.09
          \\ \hline
\end{tabularx}
\end{table*}

\begin{table*}[t!]
\centering
\small
\caption{Results on three PEFT methods based on TPP. For TPP, we used DINO as the pretext task. We highlight in \textbf{bold} the better results comparing each PEFT method with and without TPP.}
\label{tab:PEFT_dino}
\begin{tabularx}{\linewidth}{c *{3}{>{\centering\arraybackslash}X} | *{3}{>{\centering\arraybackslash}X} | *{2}{>{\centering\arraybackslash}X} | *{2}{>{\centering\arraybackslash}X}}
\hline
\multirow{2}{*}{\textbf{}} 
    & \multicolumn{3}{c|}{MHIST} 
    & \multicolumn{3}{c|}{Retinal OCT-C8} 
    & \multicolumn{2}{c|}{GlaS} 
    & \multicolumn{2}{c}{QaTa} \\ \cline{2-11} 
    & ACC & AUC & F1 
    & ACC & AUC & F1 
    & Dice & HD 
    & Dice & HD \\ \hline
Adapter 
    & 81.58 & \textbf{89.09} & 80.16 
    & 95.36 & 99.74 & 95.36 
    & \textbf{88.98} & 49.06 
    & 78.64 & 30.54 \\
Adapter w/ TPP 
    & \textbf{82.29} & 88.59 & \textbf{81.29} 
    & \textbf{96.25} & \textbf{99.75} & \textbf{96.24} 
    & 88.60 & \textbf{46.98} 
    & \textbf{79.11} & \textbf{29.27} \\ \hline
VPT 
    & 72.98 & 79.84 & 71.35 
    & 89.32 & 99.04 & 89.25 
    & 86.27 & \textbf{52.23} 
    & 77.03 & 33.43 \\
VPT w/ TPP 
    & \textbf{74.31} & \textbf{81.21} & \textbf{71.97} 
    & \textbf{91.14} & \textbf{99.24} & \textbf{91.13} 
    & \textbf{87.06} & 58.63 
    & \textbf{77.06} & \textbf{32.81} \\ \hline
SSF 
    & 82.29 & 89.29 & 80.85 
    & 88.68 & 98.93 & 88.66 
    & 87.64 & \textbf{54.49} 
    & 78.28 & \textbf{30.53} \\
SSF w/ TPP 
    & \textbf{83.32} & \textbf{90.69} & \textbf{82.04} 
    & \textbf{89.82} & \textbf{99.12} & \textbf{89.81} 
    & \textbf{88.07} & 56.43 
    & \textbf{78.61} & 30.97 \\ \hline
\end{tabularx}
\end{table*}

\begin{table*}[t!]
\centering
\small
\caption{Results of Adapter and different TPP variants with pre-training on one dataset and fine-tuning on another dataset. The pretext task is MAE and the fine-tuning framework is Adapter. The best result on each column is highlighted in \textbf{bold}.}
\label{tab:generalization}
\begin{tabularx}{\linewidth}{c |>{\centering\arraybackslash}p{1.0cm} | *{3}{>{\centering\arraybackslash}X} |*{2}{>{\centering\arraybackslash}X} |*{3}{>{\centering\arraybackslash}X} |*{2}{>{\centering\arraybackslash}X} |*{3}{>{\centering\arraybackslash}X}}
\hline
\multirow{2}{*}{\textbf{}} & \multirow{2}{=}{\centering Modality}& \multicolumn{3}{c|}{MHIST}   & \multicolumn{2}{c|}{GlaS}   & \multicolumn{3}{c|}{Chest X-ray PA}  & \multicolumn{2}{c|}{QaTa} & \multicolumn{3}{c}{Retinal OCT-C8} \\ \cline{3-15} 
                           & & ACC     & AUC    & F1     & Dice         & HD    & ACC     & AUC     & F1      & Dice         & HD    & ACC     & AUC     & F1                     \\ \hline
MHIST         & \multirow{2}{=}{\centering Path.} & \textbf{83.42} & \textbf{91.38} & \textbf{82.21} & \textbf{89.65} & 45.54          & 95.74          & 99.62          & 95.71          & 78.84          & 29.69          & 96.00          & 99.74          & 95.99          \\
GlaS           &       & 82.40          & 89.56          & 81.25          & 89.41          & \textbf{44.25} & 95.08 & 99.55 & 95.07 & 78.99          & 29.72          & 96.07          & 99.74          & 96.07          \\
\hline
Chest X-ray PA  & \multirow{2}{=}{\centering X-ray} & 79.12          & 88.16          & 78.04          & 88.51          & 50.36          & \textbf{97.38}          & \textbf{99.63}          & \textbf{97.37}          & 78.76          & 30.13          & 95.71          & \textbf{99.78} & 95.72          \\
QaTa           &       & 79.63          & 87.96          & 78.30          & 89.20          & 44.27          & 94.97          & 99.24          & 94.95          & \textbf{79.50} & \textbf{29.28} & 95.89          & 99.74          & 95.88          \\
\hline
Retinal OCT-C8 &OCT    & 81.78          & 89.32          & 80.06          & 88.32          & 47.75          & 95.19          & 99.56          & 95.15          & 78.68          & 30.23          & \textbf{96.61} & 99.77          & \textbf{96.61} \\
\hline
Adapter       &-     & 81.58          & 89.09          & 80.16          & 88.98          & 49.06          & 94.54          & 99.50          & 94.48          & 78.61          & 29.91          & 95.36          & 99.74          & 95.36         
                         \\ \hline
\end{tabularx}
\end{table*}

\begin{table*}[t]
\centering
\small
\caption{Results on different initialization strategies. We used Adapter as the baseline. Transfer means initializing target parameters by pre-training on a larger, similar dataset. Upstream means initializing target parameters by pre-training on upstream data. The best result on each column is highlighted in \textbf{bold}.}
\label{tab:initial method}
\begin{tabularx}{\linewidth}{c *{3}{>{\centering\arraybackslash}X} | *{3}{>{\centering\arraybackslash}X} | *{2}{>{\centering\arraybackslash}X} | *{2}{>{\centering\arraybackslash}X}}
\hline
\multirow{2}{*}{\textbf{}} & \multicolumn{3}{c|}{MHIST}   & \multicolumn{3}{c|}{Retinal OCT-C8}   & \multicolumn{2}{c|}{GlaS} & \multicolumn{2}{c}{QaTa} \\ \cline{2-11} 
                           & ACC     & AUC    & F1     & ACC     & AUC     & F1      & Dice         & HD        & Dice         & HD        \\ \hline
Random                    & 81.58      & 89.09      & 80.16     & 95.36      & 99.74      & 95.36     & 88.98          & 49.06          & 78.61          & 29.91          \\
Transfer                   & 82.91      & 89.51      & 81.38     & 94.61      & 99.66      & 94.62     & \textbf{89.46}          & \textbf{42.85}          & 79.13          & 29.93          \\ 
TPP w/ DINO                & 82.29      & 88.59      & 81.29     & 96.25      & 99.75      & 96.24     & 88.60          & 46.98          & 79.11          & \textbf{29.27}          \\
TPP w/ MAE                 & \textbf{83.42}      & \textbf{91.38}      & \textbf{82.21}     & \textbf{96.61}      & \textbf{99.77}      & \textbf{96.61}     & 89.41          & 44.25          & \textbf{79.50}          & 29.28          \\
Upstream                   &  82.09   & 89.59    & 79.97  & -  & -  & -  & 89.28  & 44.66  &79.34   &29.28    \\\hline
\end{tabularx}
\end{table*}

\section{Experiments}
\subsection{Datasets}
The TPP framework was evaluated on five downstream datasets: MHIST \cite{mhist}, Retinal OCT-C8 \cite{oct}, Chest X-ray PA \cite{chest}, GlaS \cite{glas}, and QaTa-COVID19-v2 (QaTa) \cite{qata}. Detailed information for these datasets is available in Table \ref{tab: dataset}.
\textbf{MHIST Dataset \cite{mhist}:} For the classification task, this dataset includes 3,152 colorectal polyp images, categorized into hyperplastic polyps (HP) and sessile serrated adenomas (SSA), with 2,175 training and 977 test images. Following the official data splits, 20\% of the training data from each category was randomly selected for validation.
\textbf{Retinal OCT-C8 Dataset \cite{oct}:} For the classification task, this dataset contains 24,000 optical coherence tomography (OCT) 2D images in eight categories: age-related macular degeneration (AMD), choroidal neovascularization (CNV), central serous retinopathy (CSR), diabetic macular edema (DME), macular hole (MH), choroidal drusen (DRUSEN), diabetic retinopathy (DR), and normal retina (NORMAL). The official data split was used for training, validation, and test sets.
\textbf{Chest X-ray PA Dataset \cite{chest}:} For the classification task, this dataset comprises 4,575 2D chest X-ray images categorized into COVID-19 infections, other pneumonia infections, and healthy controls. Due to the absence of official splits, the data was randomly divided into 70\% for training, 10\% for validation, and 20\% for testing from each category.
\textbf{GlaS Dataset \cite{glas}:} For the segmentation task, this dataset contains 165 pathological images of H$\&$E-stained colon tissue sections, labeled as malignant or benign. The official data splits were followed, and 20\% of the training data from each category was randomly selected for validation.
\textbf{QaTa-COV19v2(QaTa) Dataset \cite{qata}:} For the segmentation task, this dataset is used for segmenting COVID-19 infected regions. It consists of 7,145 training images and 2,113 test images. The official data splits were used, and 20\% of the training samples were allocated to form the validation set.


\subsection{Implementation Details}
\noindent \textbf{TPP based on MAE.}
For target parameter pre-training with MAE, we adopted the methodology detailed in ~\cite{MedCoSS, mae}. The AdamW optimizer was employed with a learning rate of 1.5e-3, a weight decay of 1.5e-2, and a batch size of 64. Pre-training was conducted for a maximum of 500 epochs for classification tasks and 1000 epochs for segmentation tasks, with the latter requiring more epochs due to smaller dataset sizes. A warm-up strategy over the initial 40 epochs linearly increased the learning rate to 1.5e-3, followed by a cosine decay to zero.

\noindent \textbf{TPP based on DINO.}
For target parameter pre-training with DINO, we followed the configuration in~\cite{dino}. The AdamW optimizer was used with a batch size of 64. The learning rate underwent a linear warm-up over the first 10 epochs, increasing to a base value determined by the linear scaling rule: lr=0.0001$\times$batch size/256, followed by a cosine decay. Similarly, weight decay followed a cosine schedule from 0.04 to 0.4. Data augmentations consistent with DINO, including color jittering, Gaussian blur, and solarization, were applied.

\noindent \textbf{Downstream Fine-tuning Stage.}
During fine-tuning, we followed the framework by MedCoSS \cite{MedCoSS}, using TPP based on MAE as the default configuration. Experiments were also conducted with TPP based on DINO. The AdamW optimizer was utilized with hyperparameters specific to each downstream task. Table \ref{tab:imple_Detail} presents the implementation details for the five downstream datasets, including loss functions, patch sizes, learning rates, batch sizes, and maximum iterations. For MHIST, a grid search for the learning rate between 0.000005 and 0.05 was performed due to the sensitivity of PEFT methods. The learning rates for full fine-tuning, linear probe, EVP, VPT, SSF, BitFit, Adapter, AdaptFormer, and DTL were set to 0.00001, 0.01, 0.00005, 0.0001, 0.001, 0.001, 0.0001, 0.001, and 0.005, respectively.

\noindent \textbf{Image Sizes of the Datasets.}
Images in MHIST and QaTa are originally $224 \times 224$ pixels. For Retinal OCT-C8, image sizes range from $384 \times 496$ to $1536 \times 496$ pixels. All images were resized to $224 \times 224$ pixels for uniformity. Similarly, Chest X-ray PA images, with original sizes from $224 \times 224$ to $5623 \times 4757$ pixels, were uniformly resized to $224 \times 224$ pixels. The GlaS dataset, initially $522 \times 775$ pixels, was resized to $512 \times 512$ pixels.

\subsection{Evaluation Metrics}
For classification tasks, we used the area under the receiver operator curve (AUC, \%), accuracy (ACC, \%), and F1 score (F1, \%) as evaluation metrics. For segmentation tasks, performance was evaluated using the Dice similarity coefficient (Dice, \%) and the 95$\%$ Hausdorff distance (HD).

\subsection{Comparisons with State-of-the-art Methods}
We compared Adapter with TPP against nine methods, including full fine-tuning, linear probe, EVP~\cite{evp}, VPT~\cite{vpt}, SSF~\cite{ssf}, BitFit~\cite{bitfit}, Adapter~\cite{adapter}, AdaptFormer~\cite{adaptformer}, and DTL~\cite{dtl}, on the MHIST, Retinal OCT-C8, Chest X-ray PA, GlaS, and QaTa datasets. MedCoSS served as the pre-trained backbone. Full fine-tuning updates all model parameters, while linear probe only updates the task-specific head. The results in Tables \ref{tab:cls} and Table \ref{tab:seg} indicate the following: (1) Among the PEFT methods, Adapter achieves the best generalization performance across all datasets. (2) Applying the proposed TPP to Adapter leads to significant performance improvements on all datasets. For instance, TPP enhances Adapter's performance on MHIST by 1.84\% (ACC), 2.29\% (AUC), and 2.05\% (F1). On GlaS, TPP improves Dice by 0.43\% and HD by 4.81 pixels. (3) Adapter with TPP outperforms full fine-tuning on MHIST, Retinal OCT-C8, Chest X-ray PA, GlaS, and QaTa, while updating approximately 3\%, 3\%, 3\%, 5\%, and 5\% of the parameters, respectively. 

For qualitative comparison, Figure. \ref{fig:visual} visualizes the segmentation results of AdaptFormer, BitFit, EVP, SSF, VPT, DTL, linear probe, Adapter, full fine-tuning, and our proposed method (Adapter with TPP) on the GlaS and QaTa datasets. These visualizations demonstrate that the segmentation results of our approach more closely match the ground truth, effectively mitigating both over- and under-segmentation. For instance, in the second row, all competing methods exhibit under-segmentation of glandular tissue, whereas our TPP method achieves more accurate segmentation.  Notably, compared to full fine-tuning, our method achieves comparable results in the first row and more accurate results in the subsequent rows.

\begin{figure*}[t]
    \centering
    \includegraphics[width=0.65\linewidth]{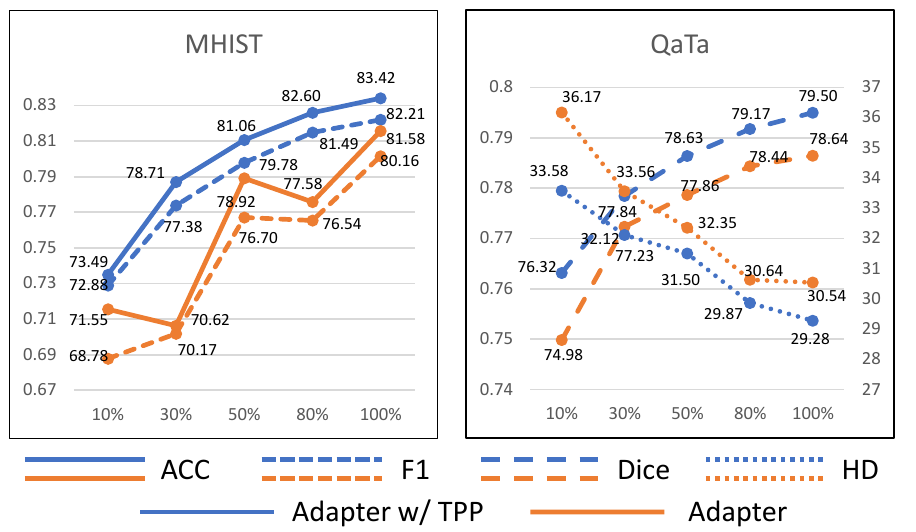} 
    \caption{Results of Adapter with and without TPP on the MHIST and QaTa with 10\%, 30\%, 50\%, 80\%, and 100\% training data.}
    \label{fig:data_ratio}
\end{figure*}

\subsection{Other PEFT Methods with TPP}
To further validate the effectiveness of TPP, we extended its application to three additional PEFT methods: EVP, VPT, and SSF, and reported their performance with and without TPP. MedCoSS was used as the pre-trained backbone. 
The results in Table \ref{tab:PEFT} confirm the robustness of our TPP approach, showing consistent positive effects across most metrics and datasets. 
For instance, the combination of EVP with TPP surpasses the performance of EVP alone across all datasets, including both classification and segmentation tasks. Specifically, on MHIST, EVP combined with TPP yields performance improvements of 0.82\% (ACC), 0.83\% (AUC), and 0.31\% (F1) compared to EVP alone. Similarly, on GlaS, adding TPP to EVP results in gains of 0.61\% in Dice and 1.04 pixels in HD over EVP.

\subsection{TPP based on DINO}
While the default pre-training pretext task for TPP is MAE, we also employed DINO \cite{dino} as the pretext task and evaluated three PEFT methods: Adapter, VPT, and SSF, as summarized in Table \ref{tab:PEFT_dino}. MedCoSS served as the pre-trained backbone. The results demonstrate that TPP, when using DINO, also enhanced fine-tuning performance, achieving superior results across nearly all metrics. For instance, Adapter with TPP achieves improvements of 0.89\% (Retinal OCT-C8 ACC), 0.88\% (Retinal OCT-C8 F1), 0.47\% (QaTa Dice), and 1.27 pixels (QaTa HD).

\section{Discussion}
\subsection{Fine-tuning with Fewer Annotations}
We evaluated the performance of Adapter, with and without TPP, on the MHIST and QaTa datasets by progressively increasing the ratios of training annotations from 10\% to 100\%, as depicted in Figure \ref{fig:data_ratio}. MedCoSS was the pre-trained backbone. As the amount of training data increased, the model's performance generally improves. Throughout this process, Adapter with TPP consistently outperforms Adapter on both the classification dataset MHIST and the segmentation dataset QaTa, demonstrating the effectiveness of TPP across varying amounts of training annotations.

\subsection{Target Parameter Pre-training on Other Data}
TPP directly utilizes the target training data for pre-training. In this section, we further investigated the performance impact of using other datasets for pre-training in a cross-validation setting. TPP with MAE was the pre-training strategy, and MedCoSS was the pre-trained backbone.
Specifically, we selected five datasets spanning three modalities for evaluation and pre-trained TPP on one dataset before fine-tuning the pre-trained model on all datasets. The baseline performance of Adapter was also recorded. The results are shown in Table \ref{tab:generalization}. 
We observed that TPP achieved superior performance when the pre-training and fine-tuning data shared identical modalities, with the best performance typically obtained when using the same dataset. Moreover, using data from a different modality for pre-training often resulted in negative effects, leading to worse performance than Adapter without TPP. For example, after pre-training adapters on the Retinal OCT-C8 dataset and fine-tuning on the GlaS dataset, we observed a 0.66\% decrease in Dice performance compared to Adapter without TPP. In summary, TPP can effectively realize its potential when pre-trained on data that is either the same as or within the same modalities as the target data.

\subsection{Initialization of Target Parameters}
We explored various initialization strategies for target parameters, using MedCoSS as the pre-trained backbone and Adapter as the PEFT strategy. These strategies include random, transfer,  TPP with DINO, TPP with MAE, and upstream. Random initialization involves initializing the target parameters with random values. Transfer learning involves initializing the target parameters by using target parameters pre-trained on a larger dataset with similar characteristics, as described in \cite{su2022transferability, pvp}. Specifically, we used the CRAG \cite{crag}, OCT2017 \cite{oct2017}, MHIST \cite{mhist}, and ChestXR \cite{ChestXR} datasets for transfer learning for the MHIST, Retinal OCT-C8, GlaS, and QaTa datasets, respectively.
Upstream pre-training involves pre-training the target parameters on upstream data. For this, we utilized MedCoSS’s upstream training data, which shares the same modality as the current task. 
Consequently, the TCGA\footnote{\url{https://portal.gdc.cancer.gov/}} dataset was employed as upstream data for MHIST and GlaS, while the MIMIC-CXR dataset was employed as upstream data for QaTa.
This experiment was not conducted for Retinal OCT-C8 due to the absence of corresponding modality data for upstream pre-training. The results, presented in Table \ref{tab:initial method}, demonstrate that pre-training-based strategies consistently outperformed random initialization, emphasizing the benefits of improved target parameter initialization. Notably, TPP with MAE achieved the best generalization performance across all datasets, highlighting its superiority.

\begin{table}[t]
\centering
\small
\caption{{Results for TPP based on MAE on MHIST and QaTa with the Path\_DINO backbone. The best result on each task is highlighted in \textbf{bold}.}}
\label{tab: new_backbone}
\begin{tabularx}{0.6\linewidth}{c *{3}{>{\centering\arraybackslash}X} @{\hspace{10pt}} |*{3}{>{\centering\arraybackslash}X}}
\hline
\multirow{2}{*}{\textbf{}} & \multicolumn{3}{c|}{MHIST}        & \multicolumn{2}{c}{QaTa} \\ \cline{2-6} 
                           & ACC & AUC & F1 & Dice   & HD    \\
\hline
Adapter        & 82.91      & 90.79     & 80.99      & 80.20      & 26.92 \\
Adapter w/ TPP  & \textbf{85.67}     & \textbf{92.84}     & \textbf{83.94}     & \textbf{80.37}      & \textbf{26.54} \\\hline
VPT           & 81.99      & 88.39      & \textbf{80.39}      & 79.62      & 29.38      \\
VPT w/ TPP     & \textbf{82.40}     & \textbf{90.02}      & 80.20      & \textbf{79.80}      & \textbf{28.82}      \\\hline
\end{tabularx}
\end{table}

\begin{table}[t]
\centering
\small
\caption{{Results for TPP based on DINO on two 3D segmentation tasks with the STU-Net backbone. The best result on each task is highlighted in \textbf{bold}.}}
\label{tab: new_architecture}
\begin{tabularx}{0.6\linewidth}{c *{3}{>{\centering\arraybackslash}X} @{\hspace{10pt}} |*{3}{>{\centering\arraybackslash}X}}
\hline
\multirow{2}{*}{\textbf{}} & \multicolumn{3}{c|}{LiTS}        & \multicolumn{3}{c}{KiTS} \\ \cline{2-7} 
                           & Mean & Organ & Tumor & Mean   & Organ & Tumor    \\
\hline
Conv-Adapter        & 77.05      & \textbf{95.27}     & 58.83      & 81.24      & \textbf{95.99}   & 66.48   \\
Conv-Adapter w/ TPP  & \textbf{77.95}     & 95.25     & \textbf{60.64}     & \textbf{82.73}      & 95.81   & \textbf{69.66}   \\\hline
\end{tabularx}
\end{table}

\begin{table}[t!]
\centering
\small
\caption{Comparison of Decoder Parameter Handling Strategies in TPP Based on MAE: Freezing, Updating, and Non-Inheritance on Chest X-ray PA and QaTa Datasets. The best result on each column is highlighted in \textbf{bold}.}
\label{tab:tpp_decoder}
\begin{tabularx}{0.6\linewidth}{c *{3}{>{\centering\arraybackslash}X}|*{2}{>{\centering\arraybackslash}X}}
\hline
\multicolumn{1}{c}{\multirow{2}{*}{\textbf{}}} & \multicolumn{3}{c|}{Chest X-ray PA}                                                                    & \multicolumn{2}{c}{QaTa}                                   \\ \cline{2-6} 
\multicolumn{1}{c}{}                           & \multicolumn{1}{c}{ACC} & \multicolumn{1}{c}{AUC} & \multicolumn{1}{c|}{F1} & \multicolumn{1}{c}{Dice} & \multicolumn{1}{c}{HD} \\ \hline
Random & 96.17 & 99.35 & 96.17 & 79.43 & 29.29 \\ 
Freeze    & \textbf{97.38} & \textbf{99.63} & \textbf{97.37} & 79.01          & 29.45          \\
Update & 95.63          & 99.59          & 95.60          & \textbf{79.50} & \textbf{29.28}
                  \\ \hline
\end{tabularx}
\end{table}

\subsection{New Backbones}
To evaluate the generalization capability of our TPP, we incorporated two new pre-trained backbones: Path\_DINO \cite{path_dino} and STU-Net \cite{stunet}. 
Path\_DINO is a ViT-based model pre-trained on pathological images using DINO \cite{dino}. We evaluated its performance with two PEFT methods: Adapter and VPT. We employed MAE as the pretext task. As shown in Table \ref{tab: new_backbone}, the results consistently demonstrate performance improvements across various datasets and metrics. Specifically, with Adapter, TPP based on MAE achieves improvements of 2.76\% (MHIST ACC), 2.05\% (MHIST AUC), 2.95\% (MHIST F1), 0.17\% (QaTa Dice), and 0.38 pixels (QaTa HD).

STU-Net is a 3D CNN-based pre-trained model trained on the TotalSegmentator dataset \cite{totalsegmentator} in a supervised manner. To accommodate CNN-based architectures, we employed Conv-Adapter \cite{convadapter} as the PEFT method and DINO as the pretext task. Additionally, we introduced two 3D medical segmentation datasets, LiTS \cite{lits} and KiTS \cite{kits19}, to evaluate the effectiveness of our TPP on CNN-based models and 3D data. As shown in Table \ref{tab: new_architecture}, Conv-Adapter with TPP achieves superior performance in tumor segmentation, while exhibiting a slight decline in organ segmentation performance compared to its counterpart without TPP. These results highlight the effectiveness and generalizability of our proposed approach.

\subsection{Decoder update strategy for TPP based on MAE} 
When using an MAE-based pre-trained model, its pre-trained decoder can be directly inherited. In this section, we investigated the impact of inheriting and freezing the pre-trained decoder during TPP. To this end, we conducted experiments on the Chest X-ray PA and QaTa datasets, with the results summarized in Table \ref{tab:tpp_decoder}. Our findings indicate that: (1) inheriting the pre-trained decoder enhances classification performance but provides only a marginal improvement for segmentation tasks; (2) after adopting the pre-trained decoder, freezing it benefits classification tasks but negatively affects segmentation performance. Based on these observations, we chose to freeze the pre-trained decoder for classification tasks while allowing updates for segmentation tasks.

\section{Conclusions}
This paper introduces a novel perspective to enhance the performance of PEFT methods. Driven by the principle that \textit{if the backbone is pre-trained, the newly introduced parameters should also be pre-trained}, we propose TPP, a plug-and-play fine-tuning framework for medical image analysis. TPP incorporates an additional pre-training stage before fine-tuning, specifically designed to pre-train the new parameters introduced by PEFT methods. The proposed framework exhibits high flexibility, supporting various pre-trained backbones, diverse pretext tasks for target parameter pre-training, and various PEFT techniques. Extensive experiments across seven datasets, covering both classification and segmentation tasks, demonstrate the consistent effectiveness of TPP. Future work will focus on designing targeted pretext strategies for target parameter pre-training, aiming to establish a unified and effective approach to fully leverage pre-training benefits for newly introduced parameters.

\bibliographystyle{cas-model2-names}

\bibliography{main}

\end{document}